\documentclass[sigconf, nonacm]{acmart}
\usepackage{multirow}
\usepackage{float}

\begin{document}

\title[COGENT]{COGENT: Continuous Graph Emulators with Neural Ordinary Differential Equations for Long-Term Physical Forecasting}


\author{Zesheng Liu}
\affiliation{%
  \institution{Lehigh University}
  \city{Bethlehem}
  \state{Pennsylvania}
  \country{USA}}
\email{zel220@lehigh.edu}

\author{Maryam Rahnemoonfar}
\authornote{Corresponding Author}
\affiliation{%
  \institution{Lehigh University}
  \city{Bethlehem}
  \state{Pennsylvania}
  \country{USA}}
\email{maryam@lehigh.edu}

\renewcommand{\shortauthors}{Liu and Rahnemoonfar}

\begin{abstract}
    Graph-based surrogate models have become increasingly important for accelerating physical simulations on irregular spatial domains, but many existing emulators are trained for one-step prediction and must be recursively rolled out to obtain long-term forecasts. This autoregressive strategy can rapidly accumulate errors, especially when the underlying system evolves over long horizons under time-varying external forcing. In this work, we present COGENT, a continuous graph emulator with Neural Ordinary Differential Equations for long-term physical forecasting on irregular geospatial meshes. COGENT encodes a finite history of system states and associated forcing fields and external forcings with a graph-based history encoder, producing node-wise context vectors that capture both local spatial interactions and temporal evolution. These context vectors initialize and condition a latent Neural Ordinary Differential Equation whose dynamics are driven by interpolated future forcings and explicit relative rollout time. By modeling the forecast trajectory as a continuous latent dynamical system, COGENT can generate predictions at arbitrary future times rather than being restricted to a fixed temporal discretization. A residual decoder maps the resulting latent trajectories back to future physical states, enabling direct multi-step forecasting without repeatedly feeding predicted states back into the model. This formulation combines graph-based spatial representation, history-conditioned latent dynamics, and continuous-time rollout in a unified framework for mesh-based physical simulation emulation. In order to stabilize training with long-horizon supervision, we also propose effective rollout-horizon sampling and a progressive rollout-horizon scheduling strategy. We evaluate COGENT on transient ice-sheet simulations generated by the Ice-sheet and Sea-level System Model, demonstrating improved long-range stability over autoregressive graph baselines. These results suggest that continuous graph Neural ODEs provide a promising methodology for scalable physical forecasting on irregular geospatial meshes, particularly in applications that require stable long-horizon predictions and the ability to query system states at arbitrary times.
\end{abstract}





\received{5 June 2026}
\received[revised]{---}
\received[accepted]{---}


\maketitle

\section{Introduction}

High-fidelity numerical simulators are central to scientific discovery and engineering design, but their computational cost often limits their use in settings that require repeated forward evaluations, such as uncertainty quantification, sensitivity analysis, inverse modeling, control, optimization, and ensemble forecasting. A learned emulator, or surrogate model, seeks to approximate the input--output behavior of a trusted numerical simulator at substantially lower computational cost, thereby enabling large-scale exploration of forcing scenarios and parameter regimes that would be impractical with the original solver alone. For time-dependent physical systems, however, emulation is not simply a static regression problem, as the model must reproduce the evolution trajectory of a dynamical process over many future steps while remaining stable under changing external forcing.

Many scientific simulators evolve physical states on irregular spatial discretizations, including finite-element meshes, coastal and hydrological graphs, and adaptive computational domains. Such outputs are naturally represented as graph-structured trajectories, where nodes correspond to spatial degrees of freedom, edges encode local geometric or mesh connectivity, static node features describe spatial or physical properties, dynamic node features represent the evolving state, and time-dependent inputs represent external forcing. Graph neural networks provide a natural emulator architecture for this setting because they operate directly on irregular domains without requiring interpolation to a Cartesian grid~\cite{pfaff2021learning,Koo_Rahnemoonfar_2025,Koo_Helheim}. Existing graph emulators, particularly single-step models that predict the next state from the current state, have demonstrated strong performance in mesh-based physical simulation and numerical ice-sheet modeling. However, because these models are typically formulated as a fixed transition from \(t\) to \(t+1\), long-horizon forecasting requires autoregressive rollout, where prediction errors can accumulate over time. Recent horizon-aware extensions move beyond single-step prediction by learning direct mappings from \(t\) to \(t+\Delta\), where \(\Delta\) represents a predefined collection of future lead times. Nevertheless, their predictions remain tied to this prescribed horizon set, which limits flexibility across simulations with different temporal resolutions or forecast ranges and may require the lead-time set to be redesigned or re-tuned for each setting.

In this work, we propose COGENT, a \textbf{Co}ntinuous \textbf{G}raph \textbf{E}mulator with \textbf{N}eural Ordinary Differential Equations for Long-\textbf{T}erm Physical Forecasting, designed for long-horizon prediction of forced graph-structured simulator trajectories. Given a fixed graph structure, static node features, a finite history of physical states and forcing fields, and a known future forcing path, COGENT predicts the future physical state at requested rollout times. The model first encodes each historical graph state using a shared graph encoder, summarizes the resulting temporal sequence with a node-wise history encoder, and uses this history context to initialize a latent state \(z_0\). This latent state is then evolved by a controlled graph Neural ODE whose dynamics are conditioned on the interpolated future forcing, static node embeddings, encoded history context, normalized rollout time, and graph connectivity. Finally, the latent trajectory is decoded as a residual correction to the last observed physical state, allowing the decoder to learn future changes rather than reconstructing the full state from scratch. This formulation combines the continuous-time representation of Neural Ordinary Differential Equations (Neural ODEs) with graph-based spatial message passing and explicit conditioning on known future forcing~\cite{Chen_2018_NODE, poli2021graphneuralordinarydifferential}.

COGENT is designed to address three limitations of existing graph emulators for long-horizon scientific forecasting. First, it replaces a finite set of discrete forecast heads with a continuous latent trajectory, allowing predictions to be queried at arbitrary rollout times. Second, it uses a non-autonomous graph ODE whose vector field is conditioned on the future forcing path throughout integration, rather than using forcing only as an input to a discrete transition model. Third, the ODE vector field receives the encoded history context and relative rollout time at every solver evaluation, enabling the latent dynamics to depend on the recent dynamical regime and on forecast position. Together with residual decoding and progressive horizon training, this design directly targets stable long rollouts without recursively feeding predicted physical states back into the model input.

COGENT is broadly applicable to numerical simulations on irregular spatial domains. In this work, we evaluate it for ice-sheet dynamic emulation, where long-horizon responses to changing environmental forcing are scientifically important yet computationally expensive to simulate. We use transient Pine Island Glacier simulations generated by the Ice-sheet and Sea-level System Model (ISSM), a finite-element ice-sheet model defined on unstructured meshes~\cite{Larour_2012, Seroussi_2014_PIGSensitivity}. Pine Island Glacier is a demanding benchmark because its dynamics are strongly influenced by fast ice flow, grounding-line retreat, and ocean-induced ice-shelf thinning~\cite{Seroussi_2014_PIGSensitivity, Joughin2021,Ian_2021}. The graph representation preserves the native ISSM mesh structure, with node states corresponding to velocity components and ice thickness, and time-dependent inputs corresponding to physical forcings.

Our experiments use scenario-level splits based on melt-rate trajectories to evaluate generalization to unseen forcing cases, rather than overlapping temporal windows from the same simulation. In a full-rollout evaluation, the first 60 simulation steps are treated as known, and the model predicts the remaining trajectory without teacher forcing. Across velocity and thickness variables, COGENT achieves the strongest whole-trajectory forecasting performance among the evaluated methods. In particular, the best COGENT configuration obtains the lowest aggregate whole-trajectory RMSE, reducing error relative to both single-step autoregressive and discrete multi-horizon graph emulators. These results suggest that history-conditioned continuous graph dynamics provide an effective inductive bias for long-horizon emulation of forced physical systems on irregular meshes.

\section{Related Work}

\subsection{Graph-Based Emulators for Ice Sheet Dynamics}

Graph neural networks (GNNs) have become a powerful model class for spatiotemporal prediction when the underlying system is governed by interactions among irregular spatial locations rather than by purely Euclidean neighborhoods. By operating on nodes and edges, GNNs naturally represent irregular spatial structure, relational dependence, and information exchange across connected locations, and have been widely used in applications such as traffic forecasting~\cite{HE2023128913,10.1007/978-3-030-59410-7_49, 10680338}, weather prediction~\cite{Xu2024, Chen_2025,keisler2022forecastingglobalweathergraph}, and cryospheric prediction of polar ice-layer~\cite{liu2026kstemitknowledgeinformedspatiotemporalefficient,Liu_ICIP,Liu_Radar}. This graph-based representation is particularly suitable for scientific emulation problems, where in traditional numerical models the physical domain is discretized as irregular mesh grids.

Early neural emulators for ice-sheet dynamics mainly followed a structured-grid paradigm, using convolutional neural networks (CNNs) or fully connected networks~\cite{Jouvet_2023,Jouvet_Cordonnier_2023, Jouvet_Cordonnier_Kim_Lüthi_Vieli_Aschwanden_2022}. Although these works demonstrated the promise of deep learning surrogates for glaciology, regular-grid neural network models are less aligned with finite-element ice-sheet simulations, where spatially varying unstructured meshes allocate finer resolution to dynamically active regions and larger elements elsewhere to efficiently represent glacier geometry and flow behavior. This motivates graph-based emulators that preserve the native mesh connectivity of numerical ice-sheet models.

Koo and Rahnemoonfar~\cite{Koo_Rahnemoonfar_2025} introduced a graph convolutional network emulator for ISSM transient simulations of Pine Island Glacier, representing mesh vertices as graph nodes and predicting node-wise ice velocity and thickness. Koo et al.~\cite{Koo_Helheim} further applied GNN emulators to Helheim Glacier calving-parameter calibration, demonstrating that GNN-based surrogates can accurately reproduce glacier evolution under varying calving stress thresholds, achieve substantial speedups over finite-element simulations, and enable efficient calibration of calving parameterizations against observed ice-front migration. More recently, Liu et al.~\cite{liu2025kangcncombiningkolmogorovarnoldnetwork} proposed KAN-GCN, which combines Kolmogorov--Arnold Networks with graph convolutions, task-specific prediction heads, and a residual one-step transition formulation to improve mesh-based ice-sheet emulation. A multi-horizon graph emulator further generalizes the standard one-step residual transition from $t\!\to\!t+1$ to $t\!\to\!t+h$, where $h$ is selected from a predefined horizon set, thereby enabling joint multi-lead supervision and mitigating error accumulation during long autoregressive rollouts~\cite{liu2026shorthistorieslongfutures}.

These studies establish GNNs as effective mesh-native surrogates for ice-sheet dynamics. However, they remain discrete-time emulators, relying on initial-state prediction, repeated one-step autoregressive rollout, or a finite set of predefined forecast horizons. As a result, long-range forecasts can suffer from error accumulation as prediction errors propagate across successive rollout steps, while horizon-specific models are restricted to a predefined set of prediction times. Instead, COGENT formulates long-range ice-sheet emulation as history-conditioned continuous-time latent evolution. By encoding recent states and forcings, driving latent dynamics with future forcing paths, and decoding predictions at arbitrary requested forecast times, COGENT provides a continuous graph emulator that mitigates autoregressive error accumulation and enables stable long-horizon forecasting on irregular geospatial meshes.

\subsection{Neural Ordinary Differential Equation}

Neural Ordinary Differential Equations parameterize the time derivative of a hidden state with a neural network and compute the resulting hidden evolution trajectory using a numerical ODE solver~\cite{Chen_2018_NODE}. This formulation provides a continuous-time alternative to a finite stack of discrete transformations and has motivated a broad class of latent continuous-time sequence models. Latent ODEs, such as ODE-RNNs\cite{Rubanova_ODE_RNNs}, extend this idea to irregularly sampled time series by inferring a latent initial condition and evolving the hidden state continuously between observation times. Controlled differential equation models further generalize this problem by allowing hidden dynamics to evolve with respect to an observed control path rather than solely from an initial condition~\cite{kidger2020neuralcde}. These models provide important foundations for continuous-time representation learning, but they are primarily designed for irregular observations, latent interpolation, or sequence modeling rather than long-horizon emulation of forced graph-structured physical trajectories.

Graph Neural ODEs extend Neural ODEs to graph-structured data by replacing the continuous vector field with a graph neural network. Poli et al.~\cite{poli2021graphneuralordinarydifferential} formalize Graph Neural ODEs as continuous-depth counterparts of graph neural networks, where the input--output transformation is defined by a continuum of message-passing layers. Continuous Graph Neural Networks further interpret graph representation learning as a continuous dynamical process over node features, with dynamics motivated by graph diffusion and neighborhood interaction~\cite{Xhonneux2020CGNN}. GRAND similarly treats graph neural networks as discretizations of an underlying diffusion process on graphs, connecting graph learning with ODE/PDE-inspired propagation~\cite{Chamberlain2021GRAND}. These methods establish a close relationship between graph message passing and continuous dynamical systems, but they mainly focus on continuous-depth graph representation learning, diffusion-style propagation, or static graph prediction tasks.

COGENT differs from these existing methods in both objective and dynamical design. Rather than learning a continuous-depth graph representation, interpolating irregular observations, or performing generic spatiotemporal forecasting, COGENT targets stable long-horizon emulation of simulator trajectories on a fixed graph under known future forcing. Its latent ODE is used as a temporal rollout operator for physical states, with the vector field conditioned throughout integration on interpolated future forcing, static node embeddings, encoded history, graph structure, and normalized rollout time. Unlike initialization-only latent ODEs, the encoded history conditions both the initial latent state and the ODE vector field. Unlike Neural CDEs~\cite{kidger2020neuralcde}, COGENT does not treat forcing as the differential control path, but feeds interpolated future forcing directly into a graph-conditioned ODE vector field. Together with residual decoding, effective rollout-horizon sampling, and progressive rollout-horizon scheduling, these choices make COGENT a task-specific continuous graph emulator for stable long-horizon forecasting rather than a standard Graph Neural ODE or Neural CDE variant.
\newcommand{\R}{\mathbb{R}}
\newcommand{\concat}{\operatorname{concat}}
\newcommand{\StaticMLP}{\ensuremath{\operatorname{StaticMLP}}}
\newcommand{\StepGNN}{\ensuremath{\operatorname{StepGNN}}}
\newcommand{\InitMLP}{\ensuremath{\operatorname{InitMLP}}}
\newcommand{\DecMLP}{\ensuremath{\operatorname{DecMLP}}}
\newcommand{\Interp}{\ensuremath{\operatorname{Interp}}}
\newcommand{\ODESolve}{\ensuremath{\operatorname{ODESolve}}}
\newcommand{\GraphNet}{\ensuremath{\operatorname{GraphNet}}}
\newcommand{\TempEnc}{\ensuremath{\operatorname{TempEnc}}}

\section{Continuous Graph Emulator with Neural Ordinary Differential Equations}
\label{COGENT}

In this work, we develop COGENT, a continuous-time graph neural emulator for long-horizon prediction of physical processes on fixed spatial domains. Given a graph $G$, static node features $x_{\mathrm{static}}$, a finite history of state variables $y_{1:H}$, a history of external forcings $u_{1:H}$, and a known future forcing path, COGENT predicts the future state $\hat{y}(t)$ at requested rollout times. COGENT follows a latent neural ordinary differential equation (Neural ODE) formulation, where observed historical states and forcing inputs are first encoded into a node-wise latent initial condition, the latent state is evolved continuously under future external forcing and graph-based spatial interactions, and the resulting latent trajectory is decoded back to physical variables.

\begin{figure*}
\centering
\includegraphics[width=1\textwidth]{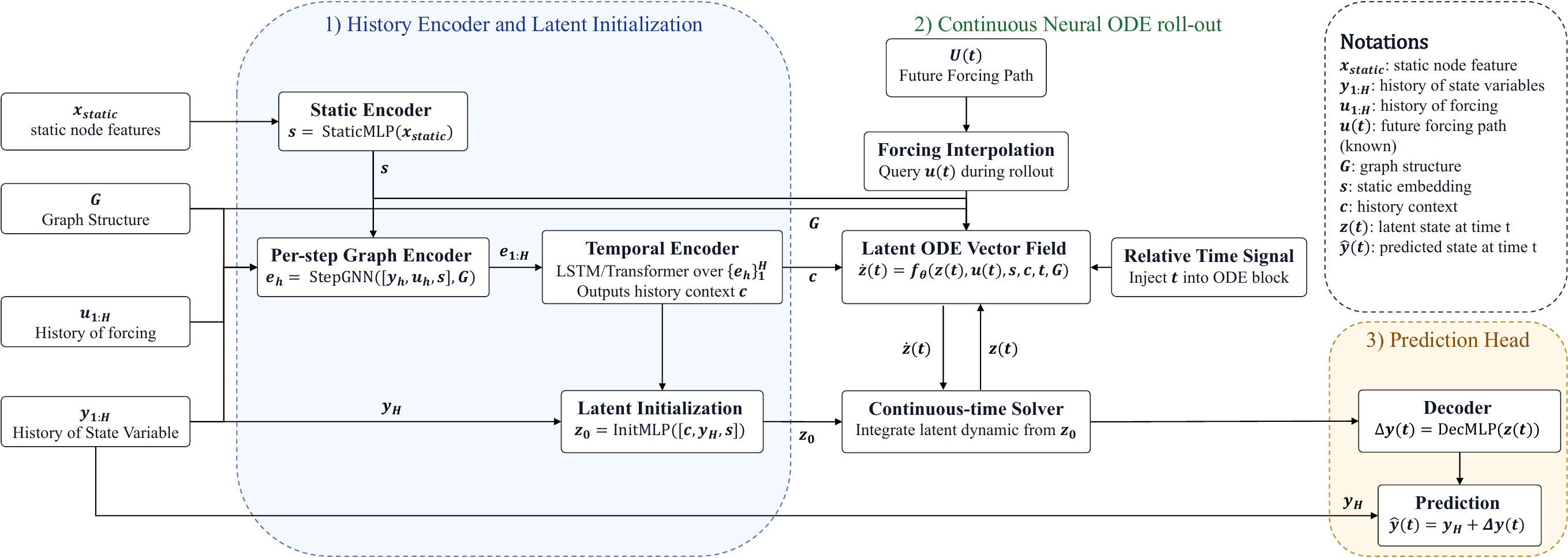}
\caption{Overview of the proposed history-conditioned latent graph Neural ODE emulator. Historical graph states and forcings are first encoded by a shared step-wise graph encoder and summarized by a temporal history encoder to obtain node-wise context vectors. The context initializes and conditions a latent continuous-time graph dynamics model driven by interpolated future forcings and relative rollout time. The latent trajectory is decoded as residual corrections to the last observed state, enabling multi-step physical forecasting on irregular spatial graphs without recursively feeding predicted states back as input.\label{fig:COGENT-Architecture}}
\Description{Overview of the proposed history-conditioned latent graph Neural ODE emulator.}
\end{figure*}  

\subsection{Problem Setup and Architecture Overview}

Let $G=(V,E)$ be a spatial graph with $N=|V|$ nodes. In our problem setup, the graph may represent an unstructured simulation mesh, a spatial adjacency graph, or a discretized geospatial domain. Each node is associated with static features $x_{\mathrm{static}} \in \R^{N\times F_x}$, which may include different kinds of global features such as coordinates, global topography, system properties, or other time-invariant attributes. At each time step, the simulator state and external forcing are represented as $y_h \in \R^{N\times F_y},u_h \in \R^{N\times F_u}$, where $h=1,\ldots,H$.
The historical inputs are denoted by $y_{1:H}=\{y_h\}_{h=1}^{H}$ and $u_{1:H}=\{u_h\}_{h=1}^{H}$, where $y_H$ is the last observed state. $H$ denotes the history length, i.e., the number of past time steps provided as input to the model. During forecasting, the future forcing path is assumed to be known and can be queried at continuous solver time $t$ through interpolation, yielding $u(t)\in\R^{N\times F_u}$. The objective is to predict $\hat{y}(t_1),\ldots,\hat{y}(t_K)$, for rollout times $0<t_1<\cdots<t_K$. Unless otherwise specified, all variables are node-wise tensors over the full graph or over the total number of nodes in a minibatch.

As shown in Figure~\ref{fig:COGENT-Architecture}, COGENT contains three main components: (i) a history encoder and latent initializer, (ii) a continuous Neural ODE rollout module, and (iii) a residual prediction head. The model first maps the raw static features to a learned static embedding, $s = \StaticMLP(x_{\mathrm{static}})$, where $s\in\R^{N\times D_s}$. For each historical step, a shared per-step graph encoder combines the step-wise state variables, forcings, and static information: $e_h = \StepGNN\big([y_h,u_h,s],G\big)$, where $e_h\in\R^{N\times D_e}$ and $h=1,\ldots,H$. This shared step-wise encoder captures spatial patterns at each historical time step and maps all history steps into a consistent feature space for temporal aggregation. The resulting sequence of graph embeddings $e_{1:H}$ is then passed to a temporal encoder to obtain a node-wise history context embedding, $c = \TempEnc(e_{1:H})$, with $c\in\R^{N\times D_c}$. This context further summarizes how the system has evolved over the input history. The initial latent state is constructed from the history context, the most recent physical state, and the static embedding: $z_0 = \InitMLP([c,y_H,s])$, where $z_0\in\R^{N\times D_z}$. Starting from $z_0$, a latent ODE evolves $z(t)$ according to $\dot{z}(t)=f_\theta\big(z(t),u(t),s,c,t,G\big)$, where $t$ denotes relative rollout time as injected into the ODE block. The ODE function is conditioned on the interpolated future forcing, static information, history context, and graph structure, allowing the latent dynamics to depend on both future drivers and historical spatial-temporal interactions. Finally, the prediction head decodes a residual update,
\begin{equation}
    \Delta y(t)=\DecMLP(z(t)),
    \qquad
    \hat{y}(t)=y_H+\Delta y(t).
    \label{eq:prediction_overview}
\end{equation}
This residual formulation directly predicts future changes from the last available state.

\subsection{History Encoder and Latent Initialization}
In COGENT, the history encoder is designed to convert the finite input history $(y_{1:H},u_{1:H})$ into the context vector $c$ that characterizes the recent dynamical regime. It separates spatial and temporal modeling: graph message passing is applied at each historical time step, and a temporal encoder then summarizes the sequence of graph-aware embeddings.

\paragraph{Static Encoder.} The static encoder transforms node attributes into a latent feature embedding,
\begin{equation}
    s = \StaticMLP(x_{\mathrm{static}}),
    \qquad s\in\R^{N\times D_s}.
    \label{eq:static_encoder}
\end{equation}
This embedding is shared by the per-step graph encoder, the latent initializer, and the latent ODE vector field. Sharing $s$ allows all major components to condition on fixed spatial and physical properties of the graph nodes.

\paragraph{Per-Step GraphSAGE Encoder.}
For each input historical step $h<H$, the encoder forms a node-wise input by concatenating the state, forcing, and static embedding:
\begin{equation}
    q_h=[y_h,u_h,s]\in\R^{N\times(F_y+F_u+D_s)}.
    \label{eq:step_input}
\end{equation}
The shared graph encoder then computes
\begin{equation}
    e_h=\StepGNN(q_h,G),
    \qquad e_h\in\R^{N\times D_e}.
    \label{eq:step_graph_encoder}
\end{equation}
The same \StepGNN{} parameters are used for all history steps, ensuring that temporal variation is captured through the sequence $e_{1:H}$ rather than through step-specific graph encoders. Inspired by previous work~\cite{liu2026kstemitknowledgeinformedspatiotemporalefficient,liu2026pactpeakawarecrossattentiongraph}, in COGENT we use GraphSAGE~\cite{Hamilton_2017_GraphSAGE} inductive framework as the \StepGNN{}.

For an unseen node $i$ with feature vector $\mathbf{x}_i$, GraphSAGE computes its updated node representation by combining its own features with the aggregated features of its neighbors:
\begin{equation}
\mathbf{x}'_i =
\mathbf{W}_{\mathrm{self}}\mathbf{x}_i
+
\mathbf{W}_{\mathrm{neigh}}
\left(
\frac{1}{|\mathcal{N}(i)|}
\sum_{j \in \mathcal{N}(i)} \mathbf{x}_j
\right),
\end{equation}
where $\mathbf{W}_{\mathrm{self}}$ and $\mathbf{W}_{\mathrm{neigh}}$ are learnable weight matrices, $\mathcal{N}(i)$ denotes the set of neighboring nodes of node $i$, and $\mathbf{x}_j$ represents the feature vector of neighbor node $j$. The mean operator aggregates information from the neighborhood, which may include nodes from different depth levels depending on the graph construction.

\paragraph{Temporal Encoder.} After per-step graph encoding, we have a sequence of spatial embeddings for each input historical step
\begin{equation}
    e_{1:H}=\{e_h\}_{h=1}^{H},
    \qquad e_{1:H}\in\R^{N\times H\times D_e}.
    \label{eq:history_sequence}
\end{equation}
The temporal encoder summarizes this sequence into the history context $c$:
\begin{equation}
    c=\TempEnc(e_{1:H}),
    \qquad c\in\R^{N\times D_c}.
    \label{eq:history_context}
\end{equation}
In this work, we apply standard temporal Transformer~\cite{Attention_2017_Vaswani} blocks over the temporal dimension as \TempEnc{}, which can be defined as
\begin{align}
    \tilde{e}_{1:H} &= e_{1:H}+p_{1:H},
    \label{eq:temporal_positional}\\
    a_{1:H} &= \operatorname{Transformer}_{\mathrm{time}}(\tilde{e}_{1:H}),
    \label{eq:temporal_transformer}\\
    c &= \operatorname{Pool}_{\mathrm{time}}(a_{1:H}),
    \label{eq:temporal_pooling}
\end{align}
where $p_{1:H}$ are temporal positional embeddings and $\operatorname{Pool}_{\mathrm{time}}$ denotes last-token pooling, mean pooling, or a learned summary-token readout. In this way, spatial dependencies are learned within each historical graph snapshot, and temporal dependencies are learned across the sequence of encoded snapshots.

\paragraph{Latent initialization.}
The initial latent state is parameterized by the node-wise history context, the last observed physical state, and the static embedding:
\begin{equation}
z_0=\InitMLP\left([c,y_H,s]\right),
\qquad z_0\in\R^{N\times D_z},
\label{eq:latent_initialization}
\end{equation}
where $[c,y_H,s]$ denotes feature-wise concatenation of the history context, last observed physical state, and static embedding. The latent state defines a learned representation space in which the continuous-time dynamics are evolved.

\subsection{Continuous Neural ODE Rollout}

The central module of COGENT is a history-conditioned latent graph Neural ODE. Given the initial latent state $z_0$, the ODE solver integrates a continuous latent trajectory $z(t)$ over the requested rollout interval. Since future forcing is provided at discrete forecast steps, the model constructs a continuous forcing signal through an interpolation operator
\begin{equation}
    u(t)=\Interp(U)(t),
    \label{eq:forcing_interpolation}
\end{equation}
where $U$ denotes the forcing sequence used to define the continuous forcing path, including the last observed forcing and the known future forcing values. This interpolation step keeps the data interface discrete while allowing the ODE solver to query the forcing at intermediate solver-internal times.

The history-conditioned latent dynamics are defined by the latent vector field
\begin{equation}
    \frac{d z(t)}{dt}=f_\theta\big(z(t),u(t),s,c,t,G\big), \qquad z(0)=z_0
    \label{eq:history_conditioned_ode}
\end{equation}
where $z(t)$ denotes the current latent state at relative rollout time $t$, $u(t)$ is the interpolated future forcing at $t$, $s$ is the static embedding, $c$ is the history context, and $G$ is the graph structure. The time variable $t$ is represented as a normalized relative-time scalar and broadcast to all nodes, providing the ODE block with explicit rollout-position information. By conditioning $f_\theta$ also on the history context $c$, the latent dynamics depend not only on the initial latent state but also on the preceding trajectory. The additional conditioning on $t$ allows the vector field to vary across forecast lead times, which improves its ability to model nonstationary temporal evolution.

In COGENT, the latent vector field $f_\theta$ is implemented as a graph neural network that directly maps the node-wise ODE input to the latent derivative. Specifically, at each relative time $t$, we form the node-wise ODE input that combines the latent state with the external forcing, static embedding, history context, and rollout time:
\begin{equation}
    r(t)=[z(t),u(t),s,c,t].
    \label{eq:ode_input}
\end{equation}
Here $t$ is broadcast to all nodes. The latent derivative is then computed directly by the graph-based vector-field block,
\begin{equation}
\dot{z}(t)
=
f_\theta\big(z(t),u(t),s,c,t,G\big)
=
\GraphNet_{\theta}(r(t),G).
\label{eq:ode_gnn}
\end{equation}
This parameterization allows the vector field to model spatial interactions during latent evolution while preserving direct access to the local latent state, forcing, static information, history context, and time signal.

Given the initial condition $z(0)=z_0$, the continuous-time solver evaluates the latent trajectory at the requested forecast times:
\begin{equation}
    z(t_1),\ldots,z(t_K)
    =\ODESolve\big(f_\theta,z_0,\{t_k\}_{k=1}^{K}\big).
    \label{eq:ode_solve}
\end{equation}
Since rollout is performed entirely in latent space, physical predictions are decoded only at the requested output times. As a result, the model avoids autoregressively feeding predicted physical states back into subsequent prediction steps.

\subsection{Residual Prediction Head}

The prediction head maps the latent trajectory to the physical state space. For each target time $t_k$, the decoder predicts a residual update
\begin{equation}
    \Delta y(t_k)=\DecMLP(z(t_k)),
    \label{eq:delta_prediction}
\end{equation}
then adds it to the last observed state:
\begin{equation}
    \hat{y}(t_k)=y_H+\Delta y(t_k).
    \label{eq:residual_prediction}
\end{equation}
This residual decoding is appropriate for physical emulation because future states are often structured perturbations of the current condition. It also stabilizes learning by allowing the decoder to focus on temporal changes rather than reconstructing the full state from scratch.
\section{Experimental Setup}

\subsection{Pine Island Glacier Simulation Dataset}
We use Pine Island Glacier (PIG), Antarctica, as the physical test case for evaluating COGENT. PIG is a challenging benchmark for continuous-time ice-sheet emulation because it is one of the most dynamically active outlet glaciers in West Antarctica~\cite{Vaughan_2001_PIGreview, Seroussi_2014_PIGSensitivity, Jenkins_2010}, and its evolution is strongly influenced by fast ice flow, grounding-line migration, and ocean-driven basal melting. The glacier has experienced substantial acceleration, thinning, and mass loss in recent decades, making it a scientifically relevant domain for testing whether a graph neural emulator can reproduce long-term ice-sheet responses under varying forcing conditions~\cite{Jacobs2011, Joughin2021, Rignot2011}.

The transient simulation trajectories are generated using the Ice-sheet and Sea-level System Model (ISSM), a finite-element ice-sheet model designed for thermomechanical simulations on unstructured meshes~\cite{Larour_2012,Seroussi_2014_PIGSensitivity,Morland_1987}. In this work, we focus on the 5km triangular mesh configuration used in prior PIG transient simulation studies~\cite{Koo_Rahnemoonfar_2025}. The mesh is designed with higher spatial resolution in dynamically sensitive regions, such as the fast-flowing trunk and grounding-line area, and coarser elements in regions with weaker spatial variability. This spatially varying discretization captures important geometric and dynamical features of the glacier while avoiding the computational cost of a uniformly high-resolution mesh across the entire domain. We construct a collection of basal-melt forcing scenarios to represent different ocean-driven melting conditions beneath the ice shelf. The prescribed basal melt rate is varied from (0) to ($70 \,\mathrm{m\,a^{-1}}$) with an increment of ($2\,\mathrm{m\,a^{-1}}$), resulting in 36 transient simulation trajectories. Each trajectory is simulated for 20 years with a monthly time step, producing 240 temporal states per melt-rate scenario. 
\subsection{Graph Construction and Scenario Splits}

Each transient simulation is represented as a graph trajectory on the native finite-element mesh. Mesh vertices are used as graph nodes, and edges are defined according to the connectivity of the triangular elements. This representation preserves the irregular spatial structure of the ISSM mesh and allows COGENT to perform message passing directly on the simulation geometry, avoiding the need to remap the outputs onto a regular grid via interpolation.

For each graph node, we construct simulation-derived static, dynamic, and geometric attributes from the transient ice-sheet simulation outputs. The static node representation consists of the mesh coordinates, \(x\) and \(y\), while the predicted state vector is defined as the ice velocity components and ice thickness \((V_x, V_y, H)\). The time-dependent forcing and auxiliary inputs include the prescribed basal melt-rate condition, surface mass balance, and floating ratio. Surface and basal elevations are also available from the simulations; however, they are not used as direct node-level covariates in this experiment. Instead, they are used to derive edge-level geometric attributes, such as local surface and bed slopes, which encode spatial variation along mesh connections. Temporal information is represented through normalized relative history and forecast times, and COGENT additionally incorporates relative rollout time within the continuous vector field used in the Neural ODE dynamics. Under this construction, each melt-rate case defines one complete graph-structured transient trajectory.

To prevent information leakage caused by overlapping temporal windows from the same simulation trajectory, we split the data by melt-rate scenario before constructing training samples. Specifically, trajectories with melt rates of \(0\), \(20\), \(40\), and \(60\,\mathrm{m\,a^{-1}}\) are used for validation, while trajectories with melt rates of \(10\), \(30\), \(50\), and \(70\,\mathrm{m\,a^{-1}}\) are held out for testing. The remaining melt-rate trajectories are used for training. This scenario-level split evaluates whether COGENT can generalize to unseen forcing cases rather than only learning from temporally adjacent windows within the same simulation.

\subsection{Temporal Windowing and Horizon Sampling}

After splitting simulations by melt-rate scenario, each complete transient trajectory is converted into temporal graph windows. For a trajectory on graph $G$, let $\left\{ \left(y_{\tau}^{\mathrm{sim}}, u_{\tau}^{\mathrm{sim}}\right) \right\}_{\tau=1}^{T}$
denote the simulated states and forcings, where $T$ is the total number of recorded simulation time steps and $\tau$ indexes the global simulation time. We use $H$ to denote the number of observed time steps provided to the model as history, and $K{\mathrm{store}}$ to denote the number of future forecast steps stored with each window for training and evaluation. A temporal window is identified by the global index $\tau_0$ of its final observed state. Given a history length $H$ and a stored future length $K{\mathrm{store}}$, the history window is defined as
\begin{equation}
\mathcal{H}{\tau_0}
=
\left\{
\left(
y_{\tau_0-H+h}^{\mathrm{sim}},
u_{\tau_0-H+h}^{\mathrm{sim}}
\right)
\right\}_{h=1}^{H}.
\label{eq:history_window}
\end{equation}
For model input, this sequence is re-indexed locally as $\mathcal{H}{\tau_0}={(y_h,u_h)}_{h=1}^{H}$, same as the notations used in Section~\ref{COGENT}.Here, $h$ denotes the local position within the extracted window rather than the global simulation index. The final history state $y_H$ therefore corresponds to $y_{\tau_0}^{\mathrm{sim}}$ and is used to initialize the latent rollout.

The same sample also stores the next $K_{\mathrm{store}}$ future states and forcings:
\begin{equation}
y(t_k) = y_{\tau_0+k}^{\mathrm{sim}},
\qquad
u(t_k) = u_{\tau_0+k}^{\mathrm{sim}},
\qquad
k=1,\ldots,K_{\mathrm{store}} .
\label{eq:sample}
\end{equation}
The relative time coordinates $t$ are anchored at the window endpoint $\tau_0$ and normalized by the characteristic timestep of the trajectory. Thus, the historical times are represented as non-positive relative offsets, while the future times correspond to positive forecast offsets. A valid endpoint must satisfy
\begin{equation}
H \leq \tau_0 \leq T-K_{\mathrm{store}},
\label{eq:K}
\end{equation}
so that both the full history block and the stored future block are available.

\subsubsection{Effective Rollout-Horizon Sampling} 

Although each window stores $K_{\mathrm{store}}$ future steps, training does not use the full stored horizon at every update. Instead, we sample an effective rollout horizon $k_{\mathrm{eff}}$ and truncate the future forcing, relative forecast times, and target states to the first $k_{\mathrm{eff}}$ steps before the model forward pass. Given a current horizon cap $K_{\mathrm{cap}}$, the sampled horizon satisfies
\begin{align}
    k_{\mathrm{eff}} \in
    \{K_{\min},K_{\min}+1,\ldots,K_{\mathrm{cap}}\},\\
    \qquad
    K_{\min}\leq K_{\mathrm{cap}}\leq K_{\mathrm{train}}^{\max}
    \leq K_{\mathrm{store}},
    \label{eq:effective_horizon_range}
\end{align}
where $K_{\min}$ is the minimum supervised horizon and $K_{\mathrm{train}}^{\max}$ is the target maximum training horizon. In our main setting, $k_{\mathrm{eff}}$ is sampled with a long-horizon-biased distribution,
\begin{equation}
    \Pr(k_{\mathrm{eff}}=k)
    =
    \frac{k}{\sum_{j=K_{\min}}^{K_{\mathrm{cap}}} j},
    \qquad
    k\in\{K_{\min},\ldots,K_{\mathrm{cap}}\}.
    \label{eq:biased_horizon_sampling}
\end{equation}
The model is then evaluated at $t_1,\ldots,t_{k_{\mathrm{eff}}}$, and the rollout loss is computed only over $\hat{y}(t_1),\ldots,\hat{y}(t_{k_{\mathrm{eff}}})$. This strategy exposes the model to variable forecast lengths while avoiding unnecessary ODE integration beyond the horizon used in the current optimization update. For distributed training, the same $k_{\mathrm{eff}}$ is used across workers for each update to keep rollout tensors synchronized.

\begin{table*}[ht]
\centering
\caption{Overall full-rollout forecasting performance. All methods are evaluated on the rollout from time steps 61 to 240, where the first 60 simulation steps are treated as known. Errors are reported as RMSE in physical units for both the whole predicted trajectory and the final forecast step. The aggregate metric summarizes the overall prediction error across \(V_x\), \(V_y\), and thickness. Lower values indicate better performance, and the best result in each column is highlighted in bold.}\label{tab:overall}
\resizebox{\textwidth}{!}{
\begin{tabular}{ccccccccc}
\toprule
\multirow{2}{*}{Method} & \multicolumn{4}{c}{Whole Trajectory} & \multicolumn{4}{c}{Final Step}   \\ \cmidrule{2-9} 
                        & $V_x$  & $V_y$  & Thickness  & Aggregated  & $V_x$ & $V_y$ & Thickness & Aggregated \\ \midrule
                    Initial-state-conditioned baseline~\cite{Koo_Rahnemoonfar_2025}    &  36.45   & 100.79    &   13.81         &    62.39         & 46.77   & 137.72   &  19.93         &   84.76         \\
                    Single Step Autoregressive Model ($t\rightarrow t+1$)~\cite{liu2026shorthistorieslongfutures}    & 70.80    & 146.24    &    21.39        &     94.62        &  105.95  & 206.79   &   32.73        &  135.47          \\
                    Multi-Horizon Emulator ($t\rightarrow t+1, t+6$)~\cite{liu2026shorthistorieslongfutures}    &  46.19   & 74.09    &   13.98         &   51.05          &  57.03  & 98.33   &    21.02       &    \textbf{66.74}        \\
                    Multi-Horizon Emulator ($t\rightarrow t+1, t+6, t+15$)~\cite{liu2026shorthistorieslongfutures}    & 50.44    & 61.96    &     \textbf{11.03}       &    46.56         &  73.22  &  92.86  &  \textbf{16.67}         &    68.95        \\
                    COGENT (Ours, $K_{\mathrm{store}}=120$)    & \textbf{28.12}    &  52.62   &   12.75         &    35.22         & \textbf{71.14}   & 115.00   &  27.98         &   79.72 \\
                    COGENT (Ours, $K_{\mathrm{store}}=150$)    & 29.21    &  \textbf{50.43}   &  11.64          &   \textbf{34.31}         & 78.01   &  \textbf{90.84}  &     25.44      &   70.67  \\ \bottomrule
\end{tabular}}
\end{table*}

\subsubsection{Progressive Rollout-Horizon Scheduling}

We further introduce progressive rollout-horizon scheduling to stabilize long-horizon Neural ODE training. Instead of exposing the model to the target maximum horizon from the beginning, we gradually increase the horizon cap used by the sampler. Let $r$ denote the training epoch, $R$ denote the number of scheduling epochs, and let $\mathcal{A}=\{\alpha_1,\ldots,\alpha_S\}$ be a non-decreasing set of warm-up fractions with $\alpha_j\in(0,1]$. During the scheduling phase, the active fraction is selected by
\begin{equation}
    j(r)
    =
    \min\left(
    \left\lfloor \frac{(r-1)S}{R} \right\rfloor + 1,
    S
    \right),
    \qquad 1\leq r\leq R.
    \label{eq:horizon_stage_index}
\end{equation}

The epoch-dependent horizon cap is then
\begin{equation}
    K_{\mathrm{cap}}^{(r)}
    =
    K_{\min}
    +
    \operatorname{round}
    \left(
    \alpha_{j(r)}
    \left(
    K_{\mathrm{train}}^{\max}-K_{\min}
    \right)
    \right),
    \label{eq:progressive_horizon_cap}
\end{equation}
with
\begin{equation}
    K_{\min}
    \leq
    K_{\mathrm{cap}}^{(r)}
    \leq
    K_{\mathrm{train}}^{\max}.
    \label{eq:progressive_horizon_constraint}
\end{equation}

After the scheduling phase, we set $K_{\mathrm{cap}}^{(r)}=K_{\mathrm{train}}^{\max}$. The effective horizon is then sampled from $\{K_{\min},\ldots,K_{\mathrm{cap}}^{(r)}\}$. This training strategy is a key component of COGENT: early epochs focus on shorter, more stable rollouts, while later epochs progressively increase the maximum supervised forecast length. The same latent vector field is optimized throughout this process, allowing the model to first learn stable short-range dynamics and then refine them under longer rollout supervision. The schedule is applied only during training; validation and test evaluation use their prescribed horizons directly.

\subsection{Training Details}

In this work, we compare COGENT against state-of-the-art graph-based emulators, including a single-step autoregressive emulator, the horizon-aware graph neural emulator~\cite{liu2026shorthistorieslongfutures}, and an initial-state-conditioned baseline~\cite{Koo_Rahnemoonfar_2025} that forecasts future states from the initial system state rather than from recent historical observations. All the networks are trained on the same GPU instance~\cite{Jetstream2_1,Jetstream2_2}, which comprises an Intel Xeon Platinum 8468 48-core CPU, 922 GB of system RAM, and 4 NVIDIA H100 GPUs, each with 80 GB of VRAM. For existing baseline methods, we follow the training settings and hyperparameters recommended in their original implementations or reported by their authors, unless otherwise specified.

For COGENT, we use a history length of $H=6$ and store $K_{\mathrm{store}}=120$ future steps for each temporal graph window to evaluate the design choices of COGENT. We also run COGENT with different $K_{\mathrm{store}}$ values to show how they affect rollout accuracy. The model is trained for 300 epochs with a batch size of 8 using the AdamW optimizer~\cite{loshchilov2018decoupled}. The initial learning rate is set to $10^{-3}$ with a weight decay of $10^{-5}$, and a cosine learning-rate scheduler is applied throughout training. We use rollout mean squared error as the training objective. Mixed-precision training is performed using bfloat16. 

As described in the previous subsection, COGENT uses progressive rollout-horizon scheduling during training. We set $k_{\min}=24$ and sample the effective horizon $k_{\mathrm{eff}}$ with a long-horizon-biased strategy, while the epoch-level maximum horizon is increased over the first 120 epochs using warm-up fractions of $0.40$, $0.55$, $0.70$, and $0.85$ relative to the gap between $k_{\min}$ and the target horizon. Gradients are clipped to a maximum norm of 1.0, with fp64 gradient-norm computation used when fp32 clipping is numerically unstable. 

For validation, we evaluate COGENT under both fixed-window forecasting and full-rollout forecasting. Model selection is based on the validation full-rollout RMSE computed in physical units. In the full-rollout setting, the first 60 simulation steps are treated as known. At the rollout start, COGENT uses only the most recent (H=6) known steps as the input history and predicts the remaining trajectory from steps 61 to 240, conditioned on the known future forcing path.

\begin{figure}[ht]
\centering
\includegraphics[width=\columnwidth]{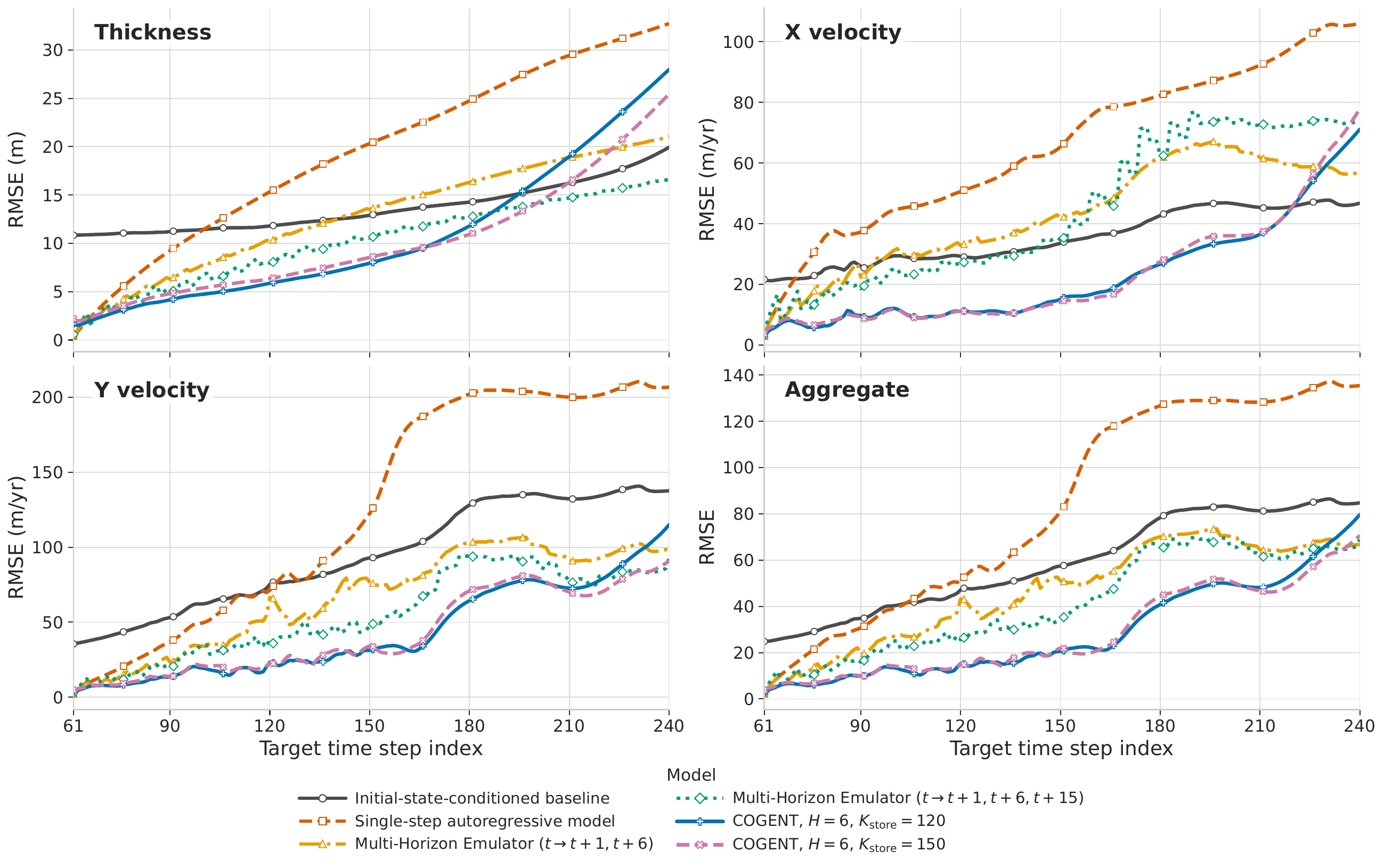}
\caption{Lead-time-wise RMSE curves across methods. The curves report RMSE at each target time step for thickness, \(V_x\), \(V_y\), and the aggregate metric over the rollout from time steps 61 to 240. This figure shows how prediction error evolves over the full rollout horizon.}\label{fig:lead_time_overall}
\Description{Lead-time-wise RMSE curves across different methods.}
\end{figure}  

\begin{figure}[ht]
\centering
\includegraphics[width=0.9\columnwidth]{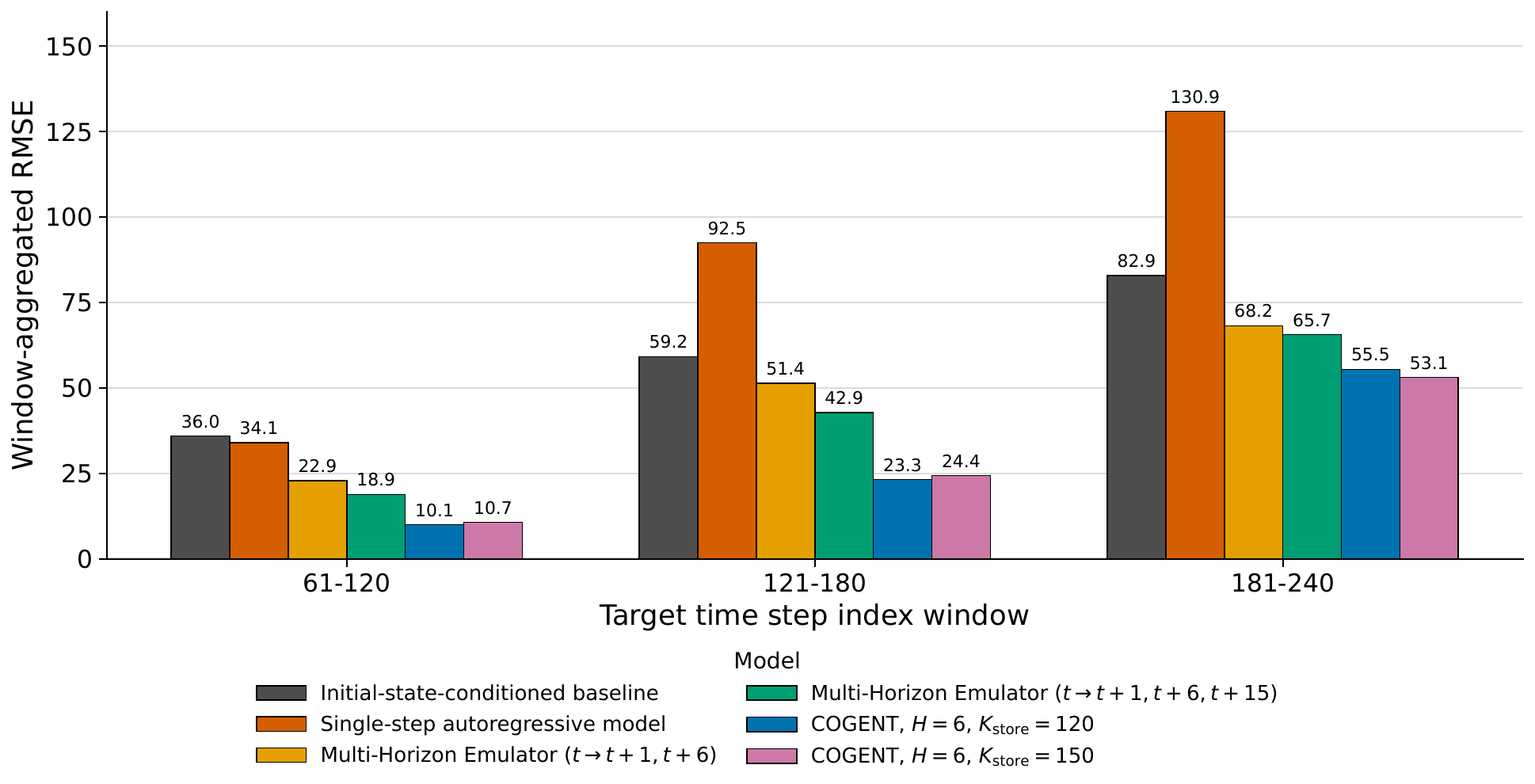}
\caption{Window-wise aggregate RMSE comparison across methods. The full rollout from time steps 61 to 240 is divided into three target-time-step windows: 61--120, 121--180, and 181--240. Bars report the aggregate RMSE averaged within each window. Lower values indicate better performance.}\label{fig:bar_plot_overall}
\Description{}
\end{figure}  
\section{Results}

\subsection{Overall Long-Horizon Forecasting Performance}

All methods are evaluated under the same full-rollout setting. The first 60 simulation steps are given, and each model predicts the remaining trajectory from time steps 61 to 240 without teacher forcing. For each predicted variable \(q \in {V_x, V_y, \mathrm{Thickness}}\), the whole-trajectory RMSE is computed over all test trajectories, graph nodes, and target time steps in \(61{:}240\). The final-step RMSE is computed only at time step 240. The aggregate RMSE is defined as the root mean square of the three variable-specific RMSE values together. We also report two rollout-level visual analyses. The lead-time-wise curves show RMSE at each target time step from 61 to 240. The window-wise bar plots report aggregate RMSE over three rollout intervals: 61--120, 121--180, and 181--240. These plots show how error accumulates across the early, middle, and late stages of rollout.

As shown in Table~\ref{tab:overall}, the single-step autoregressive model has the largest error, confirming that recursively applying a \(t\rightarrow t+1\) emulator leads to substantial error accumulation over long horizons. The multi-horizon emulators reduce this accumulation by directly supervising selected future horizons and therefore perform substantially better than the single-step model. However, COGENT achieves the strongest whole-trajectory performance. In particular, COGENT with \(K_{\mathrm{store}}=150\) obtains the lowest aggregate whole-trajectory RMSE of 34.31, reducing the error by 26.3\% compared with the best non-COGENT model and by 63.7\% compared with the single-step autoregressive model. COGENT also achieves the best whole-trajectory velocity performance, indicating that its continuous latent rollout is effective for modeling the dominant dynamical evolution over the full prediction horizon.

The final-step results are more nuanced. While COGENT with \(K_{\mathrm{store}}=150\) achieves the best final-step \(V_y\) error among all methods, the multi-horizon baselines remain competitive for endpoint metrics, especially final-step thickness and aggregate error. This behavior is expected because multi-horizon emulators are explicitly trained on a small set of predefined discrete horizons. Their performance can therefore depend strongly on the chosen horizon set, such as \(t\rightarrow t+1,t+6\) or \(t\rightarrow t+1,t+6,t+15\). In practice, this choice may need to be retuned when the simulation length, temporal resolution, or dynamical regime changes. In contrast, COGENT models the rollout through a continuous-time latent ODE conditioned on the known future forcing path, making the formulation less tied to a manually selected horizon set and easier to transfer across different rollout configurations.

Figure~\ref{fig:bar_plot_overall} further shows that COGENT reduces error throughout the rollout rather than only at a single evaluation point. Both COGENT variants achieve the lowest aggregate RMSE in all three target-time-step windows. COGENT with \(K_{\mathrm{store}}=120\) performs best in the early and middle windows, while \(K_{\mathrm{store}}=150\) gives the lowest late-window error, suggesting that increasing the stored future horizon improves late-rollout stability. The lead-time-wise curves in Fig.~\ref{fig:lead_time_overall} provide the corresponding per-variable trends and show that COGENT consistently slows error growth compared with the autoregressive and multi-horizon baselines. Beyond these discrete evaluation points, COGENT can also be queried at arbitrary relative forecast times through the Neural ODE solver. Although this flexibility is not directly quantified in the present experiments, it is an architectural advantage over discrete-horizon emulators that are trained and evaluated only at predefined time steps. Architecture-component ablations are reported in Appendix~\ref{app:ablation}, where we further examine the contribution of the main COGENT design choices.

\subsection{Sensitivity to History Window Length \texorpdfstring{$H$}{H}}

Table~\ref{tab:history_sensitivity} evaluates the effect of historical input length \(H\) while keeping the remaining model configuration fixed. The results show that increasing \(H\) does not monotonically improve forecasting accuracy, suggesting that longer historical windows do not necessarily provide more useful information for the latent dynamics. Overly short histories may provide insufficient temporal context, whereas longer histories may introduce redundant or less informative states that increase the difficulty of optimization. 

Among all tested settings, \(H=6\) achieves the best aggregate performance for both the whole predicted trajectory and the final forecast step, with errors of 35.22 and 79.72, respectively. It also obtains the lowest whole-trajectory errors for both velocity components, indicating that a six-step history provides effective temporal context for modeling the evolution of the velocity field. Compared with the second-best aggregate result, \(H=6\) reduces the whole-trajectory aggregate error from 37.17 to 35.22 and the final-step aggregate error from 82.40 to 79.72. 

Some variable-specific optima occur at other history lengths. For example, \(H=3\) achieves the lowest whole-trajectory thickness error, while \(H=8\) obtains the lowest final-step \(V_x\) and thickness errors. However, these improvements are not consistent across all predicted variables. In particular, \(H=8\) has a larger final-step \(V_y\) error than (H=6), leading to weaker aggregate performance. Since the goal of the emulator is to jointly predict velocity and thickness over long rollouts, the aggregate metric provides a more balanced criterion for selecting \(H\). Therefore, we use \(H=6\) as the best configuration for COGENT, as it provides the best overall trade-off across state variables and across trajectory-level and final-step evaluations.

\begin{table*}[ht]
\centering
\caption{Ablation study on the history window length \(H\). Errors are reported as the RMSE for the whole predicted trajectory and the final forecast step in physical units. The aggregate metric summarizes the overall prediction error across \(V_x\), \(V_y\), and thickness. Lower values indicate better performance, and the best result in each column is highlighted in bold.\label{tab:history_sensitivity}}
\resizebox{\textwidth}{!}{
\begin{tabular}{ccccccccc}
\toprule
\multirow{2}{*}{History Length $H$} & \multicolumn{4}{c}{Whole Trajectory} & \multicolumn{4}{c}{Final Step}   \\ \cmidrule{2-9} 
                                    & $V_x$  & $V_y$  & Thickness  & Aggregated  & $V_x$ & $V_y$ & Thickness & Aggregated \\ \midrule
                                   1 & 30.36    &  55.55   &   13.07         &    37.32         & 77.59   & 117.54   &  28.79         &    82.99     \\
                                   2 & 30.39    &  55.27   &   12.92         &    37.17         & 79.49   & 117.30  &  27.49         &    83.34     \\
                                   3 & 30.31    &  56.84   &   \textbf{12.39}         &    37.87         & 80.67   & 116.24   &  26.87         &    83.15    \\
                                   4 & 29.81    &  57.38   &   12.81         &    38.06         & 77.59   & 123.70   &  27.82         &    85.82      \\
                                   5 & 34.05    &  63.20   &   13.72         &    42.20         & 87.17   & 125.08   &  30.41         &   89.76       \\
                                   6 & \textbf{28.12}    &  \textbf{52.62}   &   12.75         &    \textbf{35.22}         & 71.14   & \textbf{115.00}   &  27.98         &   \textbf{79.72}       \\
                                   7 & 30.38    &  58.48   &   12.73         &    38.75         & 74.08   & 118.68   &  28.22         &    82.40     \\
                                   8 & 29.51    &  57.62   &   12.48         &    38.06         & \textbf{69.24}   & 123.41   & \textbf{26.70}          &   83.14         \\ \bottomrule
\end{tabular}}
\end{table*}

\begin{figure}[ht]
\centering
\includegraphics[width=0.95\columnwidth]{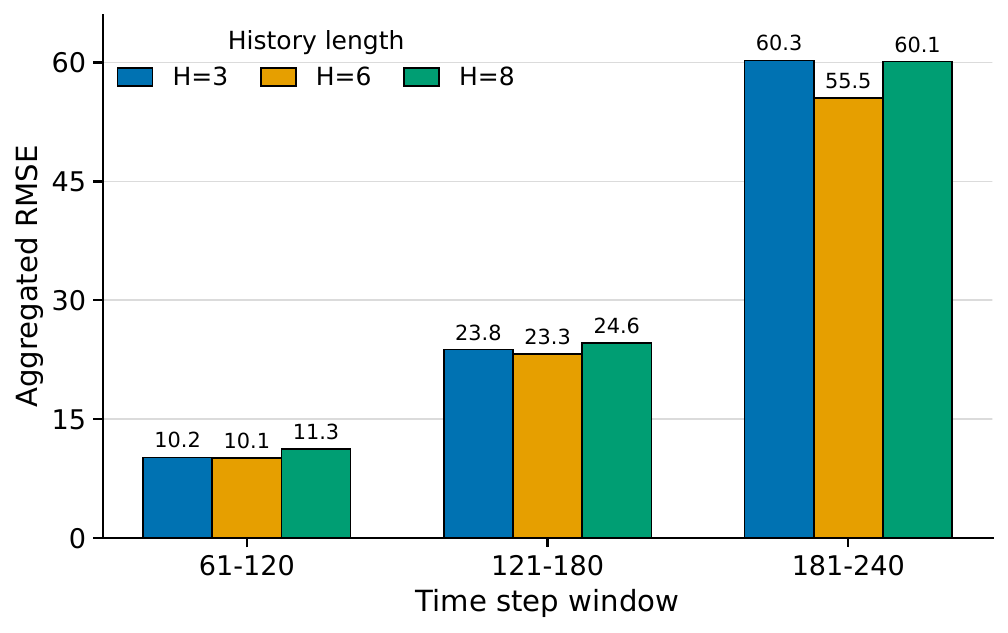}
\caption{Horizon-wise aggregate RMSE comparison for selected history lengths. The full rollout from time steps 61 to 240 is divided into three intervals: 61--120, 121--180, and 181--240. We compare \(H=3\), \(H=6\), and \(H=8\), corresponding to the best thickness-oriented setting, the best aggregate setting, and the longest tested history window, respectively.\label{fig:bar_history}}
\Description{}
\end{figure}

To further examine how the choice of \(H\) affects error accumulation over the rollout, we divide the prediction horizon from time steps 61 to 240 into three intervals: 61--120, 121--180, and 181--240. We compare \(H=3\), \(H=6\), and \(H=8\), which represent the best thickness-oriented setting, the best aggregate setting, and the longest tested history window, respectively. This horizon-wise analysis provides a more detailed view of whether the selected history length improves accuracy consistently throughout the rollout or mainly at later forecast steps. As shown in Figure~\ref{fig:bar_history}, \(H=6\) achieves the lowest aggregate RMSE in all three rollout intervals. The differences among the selected history lengths are relatively small in the early and middle intervals, but become more pronounced in the late rollout interval. In particular, \(H=6\) reduces the aggregate RMSE in the 181--240 interval from 60.3 for \(H=3\) and 60.1 for \(H=8\) to 55.5. These results further support the selection of \(H=6\), showing that it provides not only the best overall aggregate performance in Table~\ref{tab:history_sensitivity}, but also more stable error behavior throughout the full rollout horizon.

\subsection{Effect of Stored Future Horizon \texorpdfstring{$K_{\mathrm{store}}$}{K\_store}}

Table~\ref{table:K_store_ablation} evaluates the effect of the stored future horizon \(K_{\mathrm{store}}\) used during training. Increasing \(K_{\mathrm{store}}\) substantially improves full-rollout accuracy, especially when moving from short horizons to longer horizons. For example, the aggregate whole-trajectory RMSE decreases from 226.90 at \(K_{\mathrm{store}}=30\) to 35.22 at \(K_{\mathrm{store}}=120\), while the aggregate final-step RMSE decreases from 369.86 to 79.72. This indicates that training with a sufficiently long stored future horizon is critical for reducing error accumulation during long rollouts. Among the tested settings, \(K_{\mathrm{store}}=150\) achieves the lowest whole-trajectory aggregate RMSE, whereas \(K_{\mathrm{store}}=180\) achieves the best final-step aggregate RMSE. This reveals a trade-off between trajectory-average accuracy and endpoint accuracy.

\begin{table*}[ht]
\centering
\caption{Effect of the stored future horizon \(K_{\mathrm{store}}\) used during training. Errors are reported as RMSE for the full rollout from time steps 61 to 240 using both whole-trajectory and final-step evaluation. The aggregate metric summarizes the overall prediction error across \(V_x\), \(V_y\), and thickness. Lower values indicate better performance, and the best result in each column is highlighted in bold.\label{table:K_store_ablation}}
\begin{tabular}{ccccccccc}
\toprule
\multirow{2}{*}{$K_{\mathrm{store}}$} & \multicolumn{4}{c}{Whole Trajectory} & \multicolumn{4}{c}{Final Step}   \\ \cmidrule{2-9} 
                                      & $V_x$  & $V_y$  & Thickness  & Aggregated  & $V_x$ & $V_y$ & Thickness & Aggregated \\ \midrule
                                    30  & 207.89    & 326.88    &   66.14         &    226.90         &  311.00  & 550.64   &   102.31        &   369.86         \\
                                    45  & 149.79    & 272.06    &   45.67         &    181.23         &  295.23  & 534.59   &   86.62        &    356.11        \\
                                    60  & 82.11    &  152.30   &   34.01         &    101.80         &  167.27  & 316.16   &   69.93        &     210.42       \\
                                    75  & 86.17    &  124.78   &   26.28         &     88.86        &  240.10  & 295.59   &  61.15         &   222.68         \\
                                    90  &  56.59   &  87.23   &  19.09          &    61.04         & 140.64   &  195.89  &    41.45       &    141.27        \\
                                    120  & 28.12    &  52.62  &  12.75          &   35.22          & 71.14   &  115.00  &   27.98        &    79.72        \\
                                    150  & 29.21    &  \textbf{50.43}   &  \textbf{11.64}          &   \textbf{34.31}         & 78.01   &  90.84  &     25.44      &   70.67         \\
                                    180  & \textbf{27.27}    &  53.28   &  12.41          &    35.29         & \textbf{53.58}   &  \textbf{79.38}  &    \textbf{22.33}       &    \textbf{56.78}        \\ \bottomrule
\end{tabular}
\end{table*}

\begin{figure}[ht]
\centering
\includegraphics[width=0.95\columnwidth]{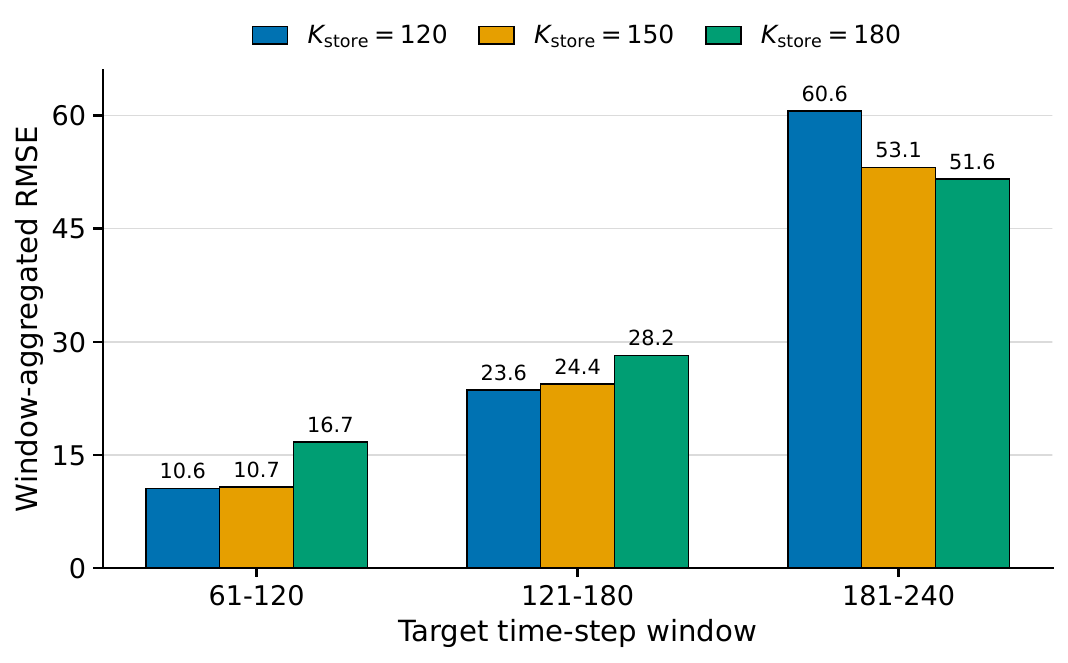}
\caption{Window-wise aggregate RMSE for different stored future horizons. The full rollout from time steps 61 to 240 is divided into three target-time-step windows: 61--120, 121--180, and 181--240. Bars compare \(K_{\mathrm{store}}=120\), \(K_{\mathrm{store}}=150\), and \(K_{\mathrm{store}}=180\).}
\label{fig:k_store_bar_plot}
\Description{}
\end{figure}

\begin{figure}[!ht]
\centering
\includegraphics[width=\columnwidth]{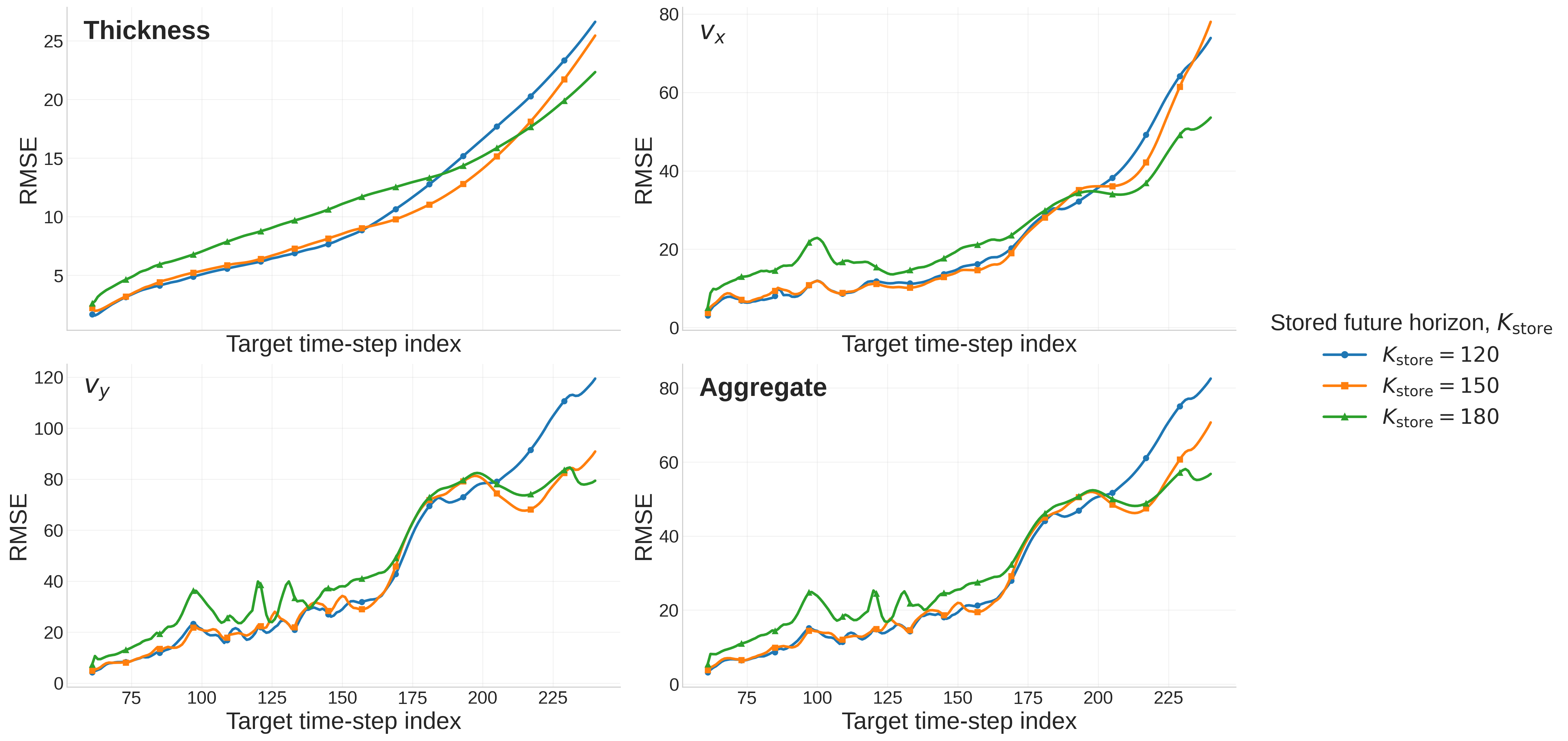}
\caption{Lead-time-wise RMSE curves for different stored future horizons. The curves report RMSE at each target time step for thickness, \(V_x\), \(V_y\), and the aggregate metric over the full rollout from time steps 61 to 240. The comparison illustrates how \(K_{\mathrm{store}}\) affects error growth across the rollout horizon.}
\label{fig:leadtime_k_store}
\Description{}
\end{figure}

Figure~\ref{fig:k_store_bar_plot} further examines this trade-off by dividing the rollout horizon into three target-time-step windows. The shorter setting \(K_{\mathrm{store}}=120\) performs best in the early and middle windows, with aggregate RMSE values of 10.6 and 23.6. However, its error increases sharply in the late window, reaching 60.6. In contrast, \(K_{\mathrm{store}}=180\) has higher error in the early and middle windows but obtains the lowest late-window RMSE of 51.6. The intermediate setting \(K_{\mathrm{store}}=150\) provides a balanced compromise, remaining close to \(K_{\mathrm{store}}=120\) in the early and middle windows while substantially reducing the late-window error. The lead-time-wise RMSE curves (Figure~\ref{fig:leadtime_k_store}) show the same pattern across thickness, velocity, and aggregate metrics. These results suggest that increasing \(K_{\mathrm{store}}\) exposes the model to longer-horizon targets during training and improves late-rollout stability, but overly emphasizing the longest horizon can reduce accuracy at shorter forecast leads.

\section{Conclusion}

We presented COGENT, a continuous graph emulator for long-horizon physical forecasting on irregular geospatial meshes. Unlike single-step autoregressive emulators that must recursively feed predictions back into the model, COGENT encodes recent state and forcing history, initializes a latent Neural ODE, and evolves the latent trajectory under known future forcing. This design combines graph-based spatial representation, history-conditioned continuous dynamics, explicit relative rollout time, and residual state prediction in a unified framework. We further introduced rollout-horizon sampling and progressive rollout-horizon scheduling to improve the stability of long-horizon training.

Experiments on transient ISSM ice-sheet simulations show that COGENT improves whole-trajectory forecasting accuracy and reduces long-range error accumulation compared with autoregressive and discrete multi-horizon graph emulators. The results also show that the stored future horizon \(K_{\mathrm{store}}\), history length \(H\), and key architectural components all affect rollout stability, with the full COGENT design providing the most balanced performance across velocity and thickness prediction. Beyond the reported discrete evaluation times, the continuous-time formulation also allows COGENT to query predictions at arbitrary relative forecast times, making it less tied to a manually selected horizon set than discrete multi-horizon emulators. These findings suggest that continuous graph Neural ODEs are a promising direction for scalable, long-term emulation of physical systems on irregular meshes.

\begin{acks}
This work is supported by NSF BIGDATA awards (IIS-1838230, IIS-2308649), NSF Leadership Class Computing awards (OAC-2139536), NSF PFI awards (2423211). This work used Jetstream2 at Indiana University through allocation CIS250588 from the Advanced Cyberinfrastructure Coordination Ecosystem: Services \& Support (ACCESS) program, which is supported by National Science Foundation grants \#2138259, \#2138286, \#2138307, \#2137603, and \#2138296.
\end{acks}

\section*{Use of Generative AI Tools.}
Generative AI tools, including ChatGPT and Codex, were used to assist with language polishing and implementation support, such as drafting, debugging, and refining portions of the data-processing, model-training, evaluation, and visualization scripts. The authors independently reviewed, modified, tested, and validated all AI-assisted code before using it in the experiments. The proposed methodology, experimental design, analysis of results, and scientific conclusions were developed and verified by the authors, who take full responsibility for the final work.

\bibliographystyle{ACM-Reference-Format}
\bibliography{reference}

\appendix



\section{Appendix: Ablation of COGENT Architectural Components}
\label{app:ablation}

\begin{table*}[!t]
\centering
\caption{Ablation study of the key architectural components in COGENT. We evaluate whether the Neural ODE vector field is conditioned on the history context \(c\), whether relative rollout time \(t\) is injected into the ODE block, and whether the prediction head uses a residual update from the last known state. Errors are reported as the RMSE for the whole predicted trajectory and the final forecast step in physical units. The aggregate metric summarizes the overall prediction error across \(V_x\), \(V_y\), and thickness. Lower values indicate better performance, and the best result in each column is highlighted in bold. \label{table:ablation-arch}}
\resizebox{\textwidth}{!}{
\begin{tabular}{ccccccccccc}
\toprule
\multirow{2}{*}{History Context in ODE} & \multirow{2}{*}{Relative Time in ODE} & \multirow{2}{*}{Residual Update} & \multicolumn{4}{c}{Whole Trajectory} & \multicolumn{4}{c}{Final Step}   \\ \cmidrule{4-11} 
                                          &                                       &                                         & $V_x$  & $V_y$  & Thickness  & Aggregated  & $V_x$ & $V_y$ & Thickness & Aggregated \\ \midrule
                                     --     &                    --                   &     --                &  33.96   & 62.84    &   15.28         &   42.17     & 76.33   &  126.28  &     38.64      &    88.07        \\
                                     \checkmark     &     --          &      --                &  36.58   & 60.75    &    14.55        &   41.80          & 94.92   & 131.36   &    34.61       &    95.68        \\
                                      --    &       \checkmark                                &      --                                   &   32.99  &   58.62  &   15.91         &   39.91          &  82.18  &  129.39  &   37.89        &    91.16        \\
                                      --    &       --                                &       \checkmark            &  35.65   &  60.99   &   14.74         &  41.67           & 95.08   & 137.90   &    34.76       &    98.77        \\
                                      \checkmark    &          \checkmark                             &    --            &  31.99   & 62.02    &  13.69          &  41.06           & 84.61   &  141.30  &    31.54       &    96.82        \\
                                      \checkmark    &          --                             &     \checkmark                                &  37.21   & 59.23    &     13.32       &     41.11        & 104.37   & 144.79   &   31.34        &     104.63       \\
                                      --    &           \checkmark             &         \checkmark             &  33.63   &  63.69   &    13.81        &   42.34          &  84.29  &  129.23  &  30.30         &    90.78        \\
                                       \checkmark   &       \checkmark                                &        \checkmark                                 & \textbf{28.12}    & \textbf{52.62}   &    \textbf{12.75}        &  \textbf{35.22}           &  \textbf{71.14}  & \textbf{115.00}   &    \textbf{27.98}      &   \textbf{79.72}         \\ \bottomrule
\end{tabular}}
\end{table*}

\begin{figure*}[!t]
    \centering
    \includegraphics[width=0.75\textwidth]{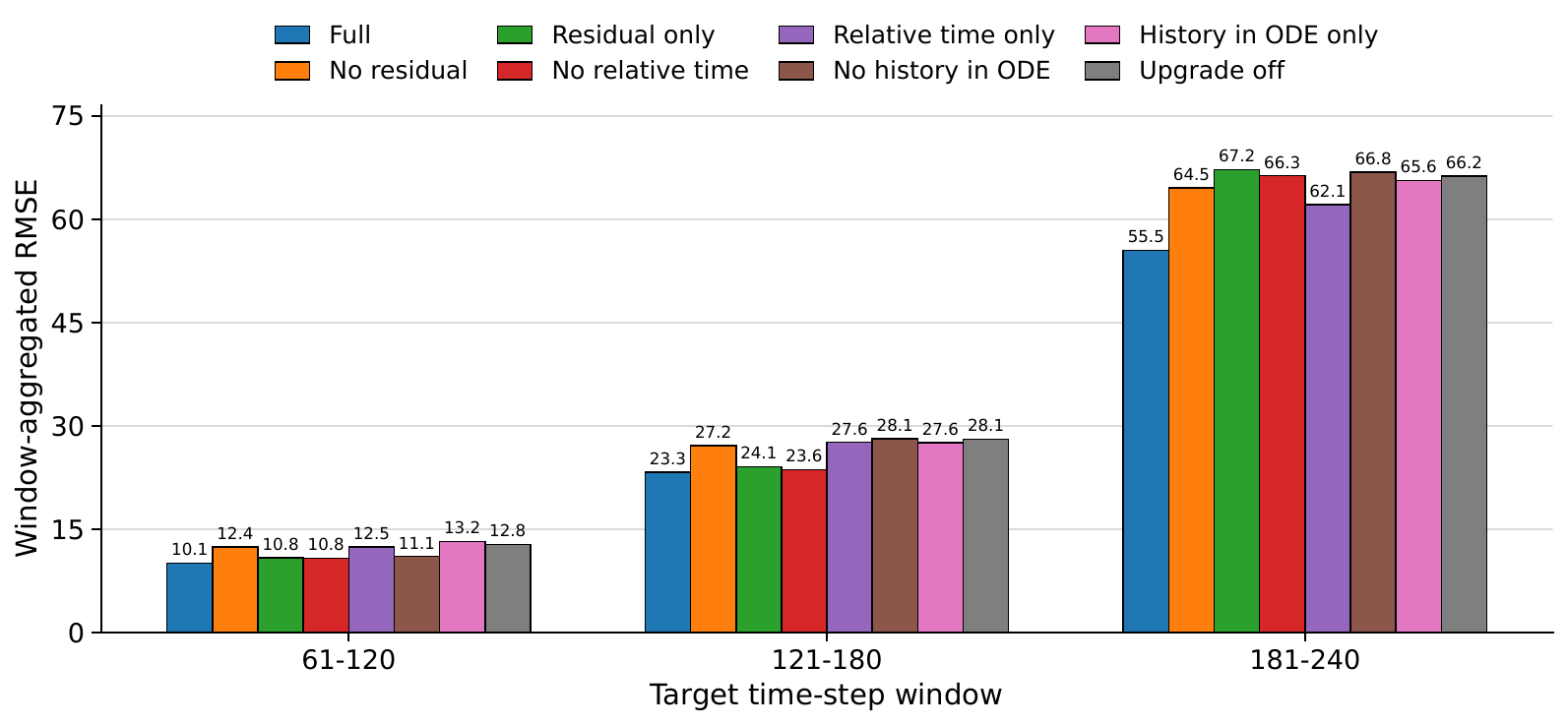}
    \caption{Window-wise ablation study of COGENT architectural components. The aggregate RMSE is reported over three rollout windows: 61--120, 121--180, and 181--240. Lower values indicate better performance.}
    \label{fig:ablation-window-rmse}
\end{figure*}

Table~\ref{table:ablation-arch} evaluates the contribution of three key architectural components in COGENT: history-context conditioning in the Neural ODE, relative-time conditioning in the Neural ODE, and residual prediction from the last known state. The full model achieves the best performance across all reported metrics, including the whole-trajectory and final-step errors for \(V_x\), \(V_y\), thickness, and the aggregate metric. Compared with the variant in which all three components are disabled, the full model reduces the aggregate whole-trajectory error from 42.17 to 35.22 and the aggregate final-step error from 88.07 to 79.72. These improvements show that the three components jointly contribute to more accurate and stable long-horizon forecasting.

Figure~\ref{fig:ablation-window-rmse} further compares the rollout behavior of different architectural variants by reporting the aggregate RMSE over three forecast windows: 61--120, 121--180, and 181--240. All variants exhibit increasing error as the rollout horizon becomes longer, confirming that late-stage prediction is the most challenging part of the task. The full COGENT model consistently achieves the lowest error in every window, with aggregate RMSE values of 10.1, 23.3, and 55.5 across the early, middle, and late rollout windows, respectively. In contrast, the variant with all three components disabled reaches 12.8, 28.1, and 66.2 over the same windows. The performance gap is especially clear in the final window, showing that the complete architecture better controls error growth during long-term rollout.

The incomplete variants further show that no single component or partial combination is sufficient to recover the full model performance. For example, using relative time alone improves the whole-trajectory aggregate error, but its final-step aggregate error remains higher than the variant without all three components. Similarly, some pairwise combinations improve thickness prediction but lead to larger velocity errors, resulting in weaker aggregate performance. The window-wise results in Figure~\ref{fig:ablation-window-rmse} show the same trend: partial variants may improve specific rollout intervals, but none consistently match the full model across the entire horizon. These results suggest that the components are complementary: the history context provides information about recent system evolution, relative time helps the Neural ODE distinguish different rollout positions, and the residual update anchors predictions to the last known physical state. Therefore, we retain all three components in the final COGENT architecture.









\end{document}